\documentclass[a4paper, 10pt, conference]{ieeeconf}      
\usepackage{FG2020}
\usepackage{graphicx}
\usepackage{amsmath}
\usepackage{amsfonts}
\usepackage{booktabs}
\usepackage{makecell}
\usepackage{balance}
\usepackage[nolist]{acronym}

\FGfinalcopy 

\IEEEoverridecommandlockouts                              
\def\FGPaperID{****} 

\title{\LARGE \bf
The FaceChannelS: Strike of the Sequences for the AffWild 2 Challenge
}

\author{\parbox{16cm}{\centering
    {\large Pablo~Barros$^1$ and Alessandra Sciutti$^1$}\\
    {\normalsize
    $^1$ Cognitive Architecture for Collaborative Technologies Unit, Istituto Italiano di Tecnologia, Genova, Italy \protect \\
    Email: \{pablo.alvesdebarros, alessandra.sciutti\}@iit.it\\

    }}
}

\begin{document}

\ifFGfinal
\thispagestyle{empty}
\pagestyle{empty}
\else
\author{Anonymous FG2020 submission\\ Paper ID \FGPaperID \\}
\pagestyle{plain}
\fi
\maketitle

\begin{abstract}

Predicting affective information from human faces became a popular task for most of the machine learning community in the past years. The development of immense and dense deep neural networks was backed by the availability of numerous labeled datasets. These models, most of the time, present state-of-the-art results in such benchmarks, but are very difficult to adapt to other scenarios. In this paper, we present one more chapter of benchmarking different versions of the FaceChannel neural network: we demonstrate how our little model can predict affective information from the facial expression on the novel AffWild2 dataset.

\end{abstract}

\section{Introduction}

Recognizing facial expressions from humans is one of the hottest topics in machine learning. The great availability of large annotated datasets helps on the attraction of the deep learning community to this effort. The results are thousands of models that perform well on several different benchmarking tasks. Most of these models are based on large deep neural networks that either use the immense feature-extraction capabilities of pre-trained models, such as the likes of VGGs and ResNet networks or are inspired by them and undergo a completely new training procedure. Our FaceChannel (FC) \cite{barros2020facechannel}, proposed recently as a light-weighted neural network for facial expression recognition, fits in the second category.

Different from traditional computer vision, however, facial expression recognition has some specific perks that makes it even more challenging. Evidence shows that humans can perceive, recognize, and commonly understand a set of `basic' affective concepts from facial expressions across cultures and around the world~\cite{Ekman1971}, however, it does not imply that we, as humans, express emotions in the same manner \cite{jack2009cultural, jack2012facial, gendron2014perceptions}. This raises a big problem, for the deep learning-based affective computing community: if each person expresses the basic emotions differently, most of the time by combining different basic concepts or even shortly transitioning between them~\cite{cavallo2018emotion}, how to adapt the large and expensive models to capture this change?

One way to address this problem, and probably the most common, is by formalizing affect in a manner that bounds the categorization ability of a computational system. This requires to choose a highly effective formalization for the task at hand~\cite{griffiths2003iii,barrett2006solving,Afzal2009}. Most of the current, and effective, solutions for automatic affect recognition are based on extreme generalization, usually employing end-to-end deep learning techniques~\cite{mehta2018facial}. Such models usually learn how to represent affective features from a large number of data samples, using strongly supervised methods~\cite{hazarika2018self,huang2019speech, Kret2013,kollias2019deep, kollias2020analysing}. 

As much of the computer vision tasks in recent times, the presence of a novel dataset that formalizes affect brings huge interest. In this regard, the AffWild2 dataset \cite{kollias2020analysing} presents a novel difficult task for emotion expression recognition. It has the largest amount of manual affective labeled data ever-present, which might help with the fine-tuning of deep learning models.

This is the case for the FaceChannel (FC) \cite{barros2020facechannel}, but in a smaller magnitude. Because it has a very light architecture, re-adapting to other datasets is not as expensive. In this paper, due to the AffWild2, we have the opportunity to demonstrate how our FaceChannel can make use of a large amount of labeled data to provide affect information from faces.

Our model presents a deep neural network inspired on the VGG-16 model, but with much fewer parameters to tune. We demonstrated recently that the FaceChannel can be easily adapted, due to having much fewer parameters to be updated during a training task, to produce competitive results in different facial expression recognition tasks.

We provide an ablation study that investigates the effect of a large amount of data on our training routine. We also describe an extension of the FaceChannel to deal with temporal data, the FaceChannelS (FC-S). In our experiments, we focus on two tasks: recognition of categorical emotions and dimensional representations, here defined as arousal and valence. We provide a series of training routines to understand the impact of a large amount of labeled data in our model. Our results demonstrate that our model still presents the fast and efficient adaptation towards new data, and provides us with a better overview of the functioning of the FaceChannel.



\section{The FaceChannel}

\begin{figure*} 
	\center{\includegraphics[width=0.9\linewidth]{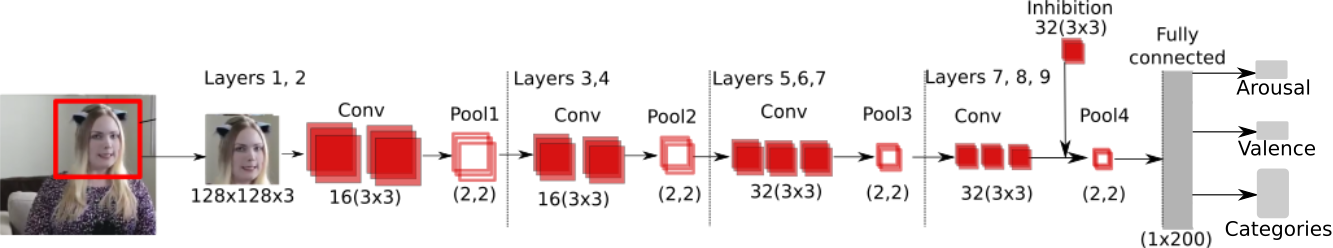}}
	\caption{Detailed architecture and parameters of The FaceChannel.}
	\label{fig:faceChannel}
\end{figure*}


In this paper, use the standard FaceChannel definition proposed and described recently \cite{barros2020facechannel}. It is implemented based on a VGG16 model~\cite{VGG2016}, but with much fewer parameters. The FaceChannel implements $10$ convolutional layers and 4 pooling layers all of them illustrated in Fig.~\ref{fig:faceChannel}. The output of the convolutional layers is fed to a fully connected layer with $500$ units, each one implementing a \textit{ReLu} activation function, which is then fed to three output layers. Each layer represent one task: predicting arousal, valence and categorical output. The arousal and valence layers are trained using a mean-squared error loss function, and the categorical classification output trained using a  using a categorical cross-entropy loss function.

\subsection{FaceChannelS}

\begin{figure*} 
	\center{\includegraphics[width=0.9\linewidth]{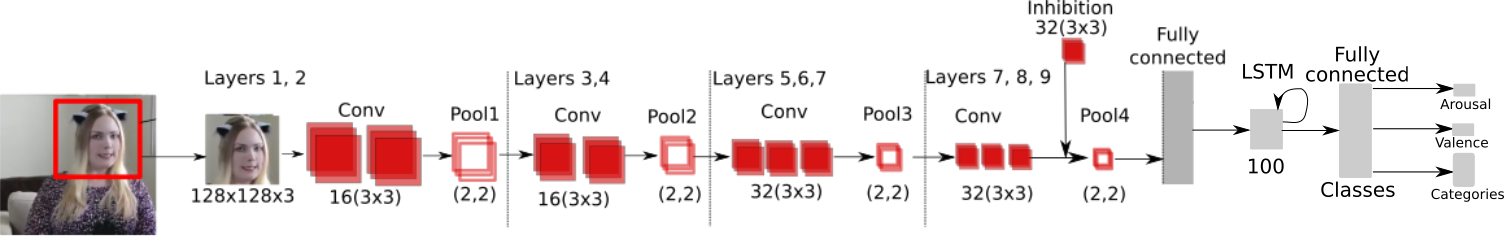}}
	\caption{Detailed architecture and parameters of The FaceChannelS.}
	\label{fig:faceChannelS}
\end{figure*}

The FaceChannelS (FC-S) follows the same structure of the FC but adds a sequence-processing layer to deal with temporal data. This layer is composed of an LSTM, with 100 units, and a dense layer with 100 units. Both of them feed to the output layer, as illustrated in Fig.~\ref{fig:faceChannelS}.

\subsection{Topology and parameters search}

As typical for most deep learning models, our FaceChannel has several hyperparameters that need to be tuned. We optimized our model to maximize the recognition accuracy using a \ac{TPE}~\cite{bergstra2011algorithms} and use the optimal training parameters throughout all of our experiments. The entire network has around $2$ \textit{million} adaptable parameters, which makes it very light-weight as compared to commonly used VGG16-based networks.

\section{Experimental Setup}

To evaluate both models (the FC and the FC-S) we perform several benchmarking experiments on the AffWild2 dataset. First we train our models using only one output: $FC_A$ and $FC-S_A$ for arousal, $FC_V$ and $FC-S_V$ for valence and $FC_E$ and $FC-S_E$ for categorical expressions. Then, we train using all three outputs: $FC$ and $FC-S$.

As the AffWild2 dataset has different labels per task, and there is a large imbalance on these labels, we proceed with a series of data pre-processing to improve the learning of our models. 

\subsection{Data Pre-processing}

We focus on analyzing our models on two tasks provided by the AffWild2 dataset: categorical classification and dimensional (arousal and valence) prediction. All the samples from the AffWild2 training set have both labels, but are extremely imbalanced, as illustrated by Figure X.

\begin{figure} 
	\center{\includegraphics[width=1\linewidth]{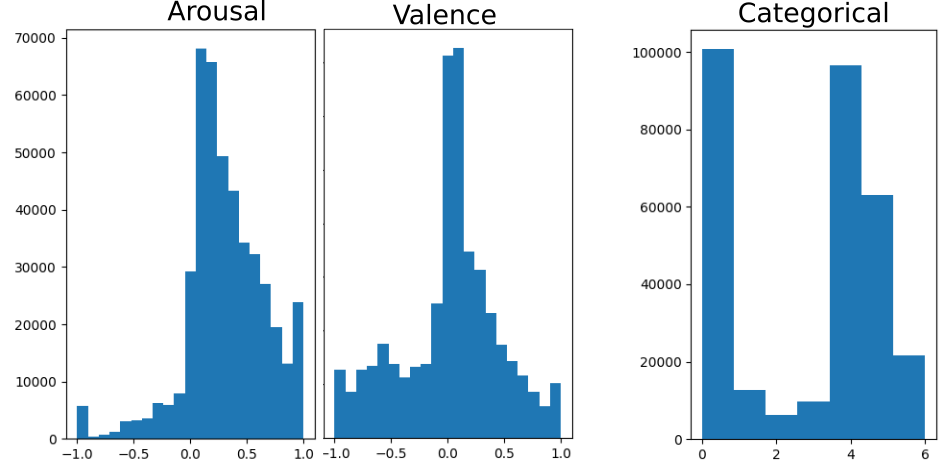}}
	\caption{Annotation distributions for the dimensional (arousal and valence) and categorical labels present in the AffWild2 dataset.}
	\label{fig:dataDistribution}
	\end{figure}

The first pre-processing we do is to extract a congruent subsample of the dataset. As each sample has both categorical and dimensional labels, it is helpful to identify which samples are ambiguous or incoherent and remove them from the dataset. We follow the subsampling proposed by Kuhnke et al. \cite{kuhnke2020two} that removes every sample that:

\begin{itemize}
    \item Invalid valence or arousal.
    \item Happy categorical expressions with negative valence.
    \item Sad categorical expressions with positive valence.
    \item Neutral categorical expressions with valence and arousal higher than 0.5.
       
\end{itemize}

After cleaning the dataset of the incoherent samples, we perform a data-augmentation routine to provide a better balanced training. For the categorical labels, we include augmented images from the same class until all the classes have the same number of samples. For the dimensional task, we first bin the samples into 21 categories, and provide with the same technique. This improved drastically the generalization capabilities of our models.

\subsection{Training Parameters}

We then proceed to use the cropped and aligned faces in our training routines. We maintain the same original dimensions of (120,120,3) pixels. For the FC-S, we provide a sequence with 10 images, having an input vector of dimension (10, 120,120,3). For all our experiments, we maintain a batch size of 1024 and trained all the networks using an RTX Quadro 4000 GPU.

The FC model was trained from the scratch for both tasks, which provided our best results. The FC-S model used the pre-trained FC model as basis, adding the sequence processing layer after the dense layer of the FC.

\subsection{Metrics}

To measure the performance of the FC and FC-S we use the following metrics: accuracy and F1-Score  to recognize categorical emotion expressions, and the CCC~\cite{Lawrence1989} between the outputs of the models and the true label to recognize arousal and valence. The \ac{CCC} is computed as:

\begin{equation}
CCC = \frac{2 \rho \sigma_x \sigma_y}{\sigma_{x}^2 + \sigma_{y}^2 + (\mu_x - \mu_y)^2}
\label{eq:ccc}
\end{equation}

\noindent where $\rho$ is the Pearson's Correlation Coefficient between model prediction labels and the annotations, $\mu_x$, and $\mu_y$ denote the mean for model predictions and the annotations and $\sigma_{x}^2$ and $\sigma_{y}^2$ are the corresponding variances. 

Both metrics are by the experimental protocols defined by each of the individual dataset authors.

\section{Results}

\begin{table}[b]
    \caption{\acf{CCC}, for arousal and valence, and the Categorical Accuracy and F1-Score when evaluating the FaceChannel(FC) and FaceChannel-S(FC-S) with the validation set of the AffWIld2 Dataset in different configurations: training on individual outputs ($FC_A$ and $FC-S_A$ for arousal, $FC_V$ and $FC-S_V$ for valence and $FC_E$ and $FC-S_E$ for categorical expressions) and joint training ($FC$ and $FC-S$). }
    \center 
    \setlength\tabcolsep{4pt}
    \footnotesize{
    \begin{tabular}{ c |  c | c | c | c }\toprule
    \textbf{Model}                          & \textbf{Arousal}  & \textbf{Valence} & \textbf{F1-Score} & \textbf{Accuracy}   \\\midrule
     \multicolumn{5}{c}{\textbf{Frame}}\\\hline
        
        $FC_A$   & 0.38  & -  &- &-  \\
        $FC_V$   &- & 0.12  & -& - \\
        $FC_E$   &- &-  & 0.31 & 0.34  \\
        $FC$     & 0.50         & 0.25  & 0.49 & 0.55  \\\\\hline
    \multicolumn{5}{c}{\textbf{Sequence}}\\\hline
        $FC-S_A$ & 0.40 & - & -  &- \\
        $FC-S_V$ &- & 0.15 & - & - \\
        $FC-S_E$ &- & - &  0.38 & 0.41 \\
        $FC-S$ & 0.53 & 0.27 &  0.52 & 0.57 \\    
        
 \\\bottomrule
    \end{tabular} 
\label{tab:AllExperiments}
}
\end{table}

All of our results are reported using the validation set of the AffWild2 dataset, in Table \ref{tab:AllExperiments}. When trained for individual tasks, our model provides the worst results of all our experiments. Using a joint training, the FC model provides a good initial estimation for both dimensional and categorical tasks. In all cases, the models present a lower valence CCC, which needs to be investigated further.

The FC-S allows the processing of temporal data, which allows the model to capture better the facial expression changes within the same video. It presents improved results when compared to FC.

\section{Conclusions}

In this paper, we present our investigation on training and validating the FaceChannel (FC) on the AffWild2 dataset. We also propose an extension of the FC capable of processing temporal faces (the FC-S) by adding a layer to process sequential data.

We perform a series of experiments to demonstrate the ability of the networks to recognize categorical and dimensional facial expressions. To guarantee the reproducibility and dissemination of our model, we have made it fully available on GitHub\footnote{https://github.com/pablovin/AffectiveMemoryFramework}.

In the future, we plan to extend the application of our model in real-world scenarios, in particular the ones involving social robots. We also believe that extending the feature extraction capabilities of our model to deal with more facial recognition tasks would be an interesting direction for our research.



\balance

\bibliographystyle{ieeetr}
\bibliography{bib}

\begin{acronym}
\acro{CCC}{Concordance Correlation Coefficient}
\acro{FER}{Facial Expression Recognition}
\acro{LSTM}{Long Short-Term Memory}
\acro{OMG-Emotion}{One Minute Gradual Emotion Recognition}
\acro{TPE}{Tree-structured Parzen Estimator}
\end{acronym}
\end{document}